\documentclass[runningheads]{llncs}
\usepackage{graphicx}

\usepackage{tikz}
\usepackage{comment}
\usepackage{amsmath,amssymb} 
\usepackage{color}

\newsavebox\CBox
\def\textBF#1{\sbox\CBox{#1}\resizebox{\wd\CBox}{\ht\CBox}{\textbf{#1}}}

\newcommand{\eg}{{\emph{e.g.}}}
\newcommand{\ie}{{\emph{i.e.}}}

\newcommand{\etc}{{\emph{etc}}}
\definecolor{grayhighlight}{RGB}{213,229,255}

\begin{document}
\pagestyle{headings}
\mainmatter
\def\ECCVSubNumber{100}  

\title{Efficient Single-Image Depth Estimation on Mobile Devices, Mobile AI \& AIM 2022 Challenge: Report} 

\titlerunning{ECCV-22 submission ID \ECCVSubNumber}
\authorrunning{ECCV-22 submission ID \ECCVSubNumber}
\author{Anonymous ECCV submission}
\institute{Paper ID \ECCVSubNumber}

\titlerunning{Efficient Single-Image Depth Estimation on Mobile Devices}
%
\author{Andrey Ignatov \and Grigory Malivenko \and Radu Timofte \and
Lukasz Treszczotko \and Xin Chang \and Piotr Ksiazek \and Michal Lopuszynski \and Maciej Pioro \and Rafal Rudnicki \and Maciej Smyl \and Yujie Ma \and
Zhenyu Li \and Zehui Chen \and Jialei Xu \and Xianming Liu \and Junjun Jiang \and
XueChao Shi \and Difan Xu \and Yanan Li \and Xiaotao Wang \and Lei Lei \and
Ziyu Zhang \and Yicheng Wang \and Zilong Huang \and Guozhong Luo \and Gang Yu \and Bin Fu \and
Jiaqi Li \and Yiran Wang \and Zihao Huang \and Zhiguo Cao \and
Marcos V. Conde \and Denis Sapozhnikov \and
Byeong Hyun Lee \and Dongwon Park \and Seongmin Hong \and Joonhee Lee \and Seunggyu Lee \and Se Young Chun $^*$
}

\institute{}
\authorrunning{A. Ignatov, G. Malivenko, R. Timofte et al.}
\maketitle

\begin{abstract}

Various depth estimation models are now widely used on many mobile and IoT devices for image segmentation, bokeh effect rendering, object tracking and many other mobile tasks. Thus, it is very crucial to have efficient and accurate depth estimation models that can run fast on low-power mobile chipsets. In this Mobile AI challenge, the target was to develop deep learning-based single image depth estimation solutions that can show a real-time performance on IoT platforms and smartphones. For this, the participants used a large-scale RGB-to-depth dataset that was collected with the ZED stereo camera capable to generated depth maps for objects located at up to 50 meters. The runtime of all models was evaluated on the Raspberry Pi 4 platform, where the developed solutions were able to generate VGA resolution depth maps at up to 27 FPS while achieving high fidelity results. All models developed in the challenge are also compatible with any Android or Linux-based mobile devices, their detailed description is provided in this paper.

\end{abstract}

{\let\thefootnote\relax\footnotetext{%
\hspace{-5mm}$^*$
Andrey Ignatov, Grigory Malivenko and Radu Timofte are the Mobile AI \& AIM 2022 challenge organizers \textit{(andrey@vision.ee.ethz.ch, grigory.malivenko@gmail.com, radu.timofte@uni-wuerzburg.de)}. The other authors participated in the challenge. Appendix \ref{sec:apd:team} contains the authors' team names and affiliations. \vspace{2mm} \\ Mobile AI 2022 Workshop website: \\ \url{https://ai-benchmark.com/workshops/mai/2022/}
}}

\section{Introduction}

\begin{figure*}[t!]
\centering
\setlength{\tabcolsep}{1pt}
\resizebox{\linewidth}{!}
{
\includegraphics[width=0.5\linewidth]{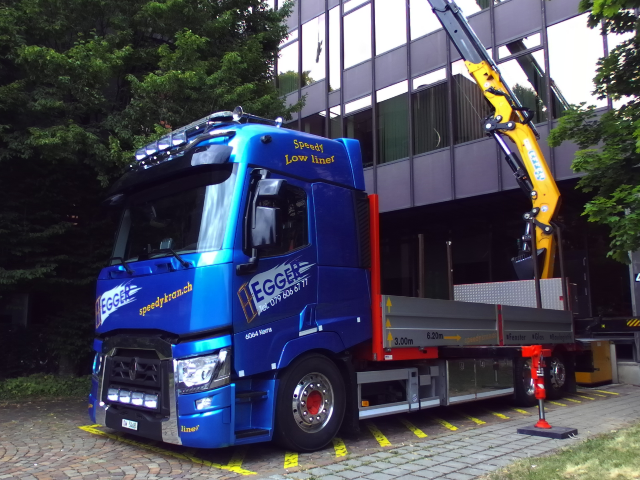} \hspace{2mm}
\includegraphics[width=0.5\linewidth]{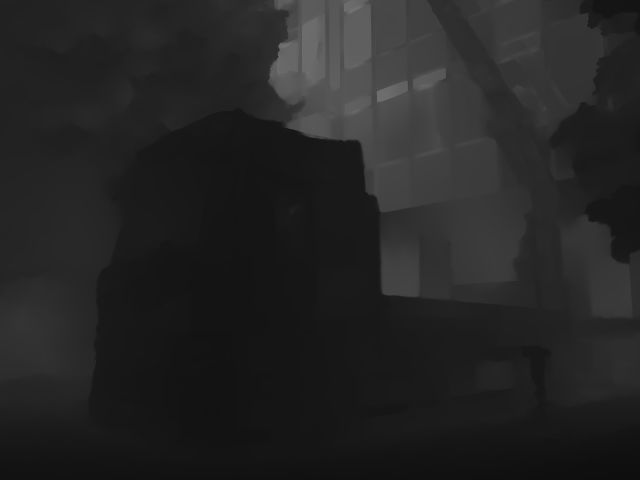}
}
\vspace{-1.2mm}
\caption{The original RGB image and the corresponding depth map obtained with the ZED 3D camera.}
\label{fig:example_photos}
\vspace{-1.2mm}
\end{figure*}

Single-image depth estimation is an important vision task. There is an ever-increasing demand for fast and efficient single-image depth estimation solutions that can run on mobile devices equipped with low-power hardware. This demand is driven by a wide range of depth-guided problems arising in mixed and augmented reality, autonomous driving, human-computer interaction, image segmentation, bokeh effect synthesis tasks.
The depth estimation research from the past decade produced multiple accurate deep learning-based solutions~\cite{li2018megadepth,godard2019digging,eigen2014depth,liu2015deep,liu2015learning,laina2016deeper,garg2016unsupervised,chen2016single}. However, these solutions are generally focusing on high fidelity results and are lacking the computational efficiency, thus not meeting various mobile-related constraints, which are key for image processing tasks~\cite{ignatov2017dslr,ignatov2018wespe,ignatov2020replacing} on mobile devices. Therefore, most of the current state-of-the-art solutions have high computational and memory requirements even for processing low-resolution input images and are incompatible with the scarcity of resources found on the mobile hardware. In this challenge, we are using a large-scale depth estimation dataset and target the development of efficient solutions capable to meet hardware constraints like the ones found on the Rasberry Pi 4 platform. This is the second installment of this challenge. The previous edition was in conjunction with Mobile AI 2021 CVPR workshop~\cite{ignatov2021fastDepth}.

The deployment of AI-based solutions on portable devices usually requires an efficient model design based on a good understanding of the mobile processing units (\eg CPUs, NPUs, GPUs, DSP) and their hardware particularities, including their memory constraints. We refer to~\cite{ignatov2019ai,ignatov2018ai} for an extensive overview of mobile AI acceleration hardware, its particularities and performance. As shown in these works, the latest generations of mobile NPUs are reaching the performance of older-generation mid-range desktop GPUs. Nevertheless, a straightforward deployment of neural networks-based solutions on mobile devices is impeded by (i) a limited memory (\ie, restricted amount of RAM) and
(ii) a limited or lacking support of many common deep learning operators and layers. These impeding factors make the processing of high resolution inputs impossible with the standard NN models and require a careful adaptation or re-design to the constraints of mobile AI hardware. Such optimizations can employ a combination of various model techniques such as 16-bit / 8-bit~\cite{chiang2020deploying,jain2019trained,jacob2018quantization,yang2019quantization} and low-bit~\cite{cai2020zeroq,uhlich2019mixed,ignatov2020controlling,liu2018bi} quantization, network pruning and compression~\cite{chiang2020deploying,ignatov2020rendering,li2019learning,liu2019metapruning,obukhov2020t}, device- or NPU-specific adaptations, platform-aware neural architecture search~\cite{howard2019searching,tan2019mnasnet,wu2019fbnet,wan2020fbnetv2}, \etc.

The majority of competitions aimed at efficient deep learning models use standard desktop hardware for evaluating the solutions, thus the obtained models rarely show acceptable results when running on real mobile hardware with many specific constraints. In this \textit{Mobile AI challenge}, we take a radically different approach and propose the participants to develop and evaluate their models directly on mobile devices. The goal of this competition is to design a fast and performant deep learning-based solution for a single-image depth estimation problem. For this, the participants were provided with a large-scale training dataset containing over 8K RGB-depth image pairs captured using the ZED stereo camera. The efficiency of the proposed solutions was evaluated on the popular Raspberry Pi 4 ARM-based single-board computer widely used for many machine learning IoT projects. The overall score of each submission was computed based on its fidelity (si-RMSE) and runtime results, thus balancing between the depth map reconstruction quality and the computational efficiency of the model. All solutions developed in this challenge are fully compatible with the TensorFlow Lite framework~\cite{TensorFlowLite2021}, thus can be efficiently executed on various Linux and Android-based IoT platforms, smartphones and edge devices.

\smallskip

This challenge is a part of the \textit{Mobile AI \& AIM 2022 Workshops and Challenges} consisting of the following competitions:

\small

\begin{itemize}
\item Efficient Single-Image Depth Estimation on Mobile Devices
\item Learned Smartphone ISP on Mobile GPUs~\cite{ignatov2022maiisp}
\item Power Efficient Video Super-Resolution on Mobile NPUs~\cite{ignatov2022maivideosr}
\item Quantized Image Super-Resolution on Mobile NPUs~\cite{ignatov2022maisuperres}
\item Realistic Bokeh Effect Rendering on Mobile GPUs~\cite{ignatov2022maibokeh}
\item Super-Resolution of Compressed Image and Video~\cite{yang2022aim}
\item Reversed Image Signal Processing and RAW Reconstruction~\cite{conde2022aim}
\item Instagram Filter Removal~\cite{kinli2022aim}
\end{itemize}

\normalsize

\noindent The results and solutions obtained in the previous \textit{MAI 2021 Challenges} are described in our last year papers:

\small

\begin{itemize}
\item Single-Image Depth Estimation on Mobile Devices~\cite{ignatov2021fastDepth}
\item Learned Smartphone ISP on Mobile NPUs~\cite{ignatov2021learned}
\item Real Image Denoising on Mobile GPUs~\cite{ignatov2021fastDenoising}
\item Quantized Image Super-Resolution on Mobile NPUs~\cite{ignatov2021real}
\item Real-Time Video Super-Resolution on Mobile GPUs~\cite{romero2021real}
\item Quantized Camera Scene Detection on Smartphones~\cite{ignatov2021fastSceneDetection}
\end{itemize}

\normalsize


\begin{figure*}[t!]
\centering
\setlength{\tabcolsep}{1pt}
\resizebox{0.96\linewidth}{!}
{
\includegraphics[width=1.0\linewidth]{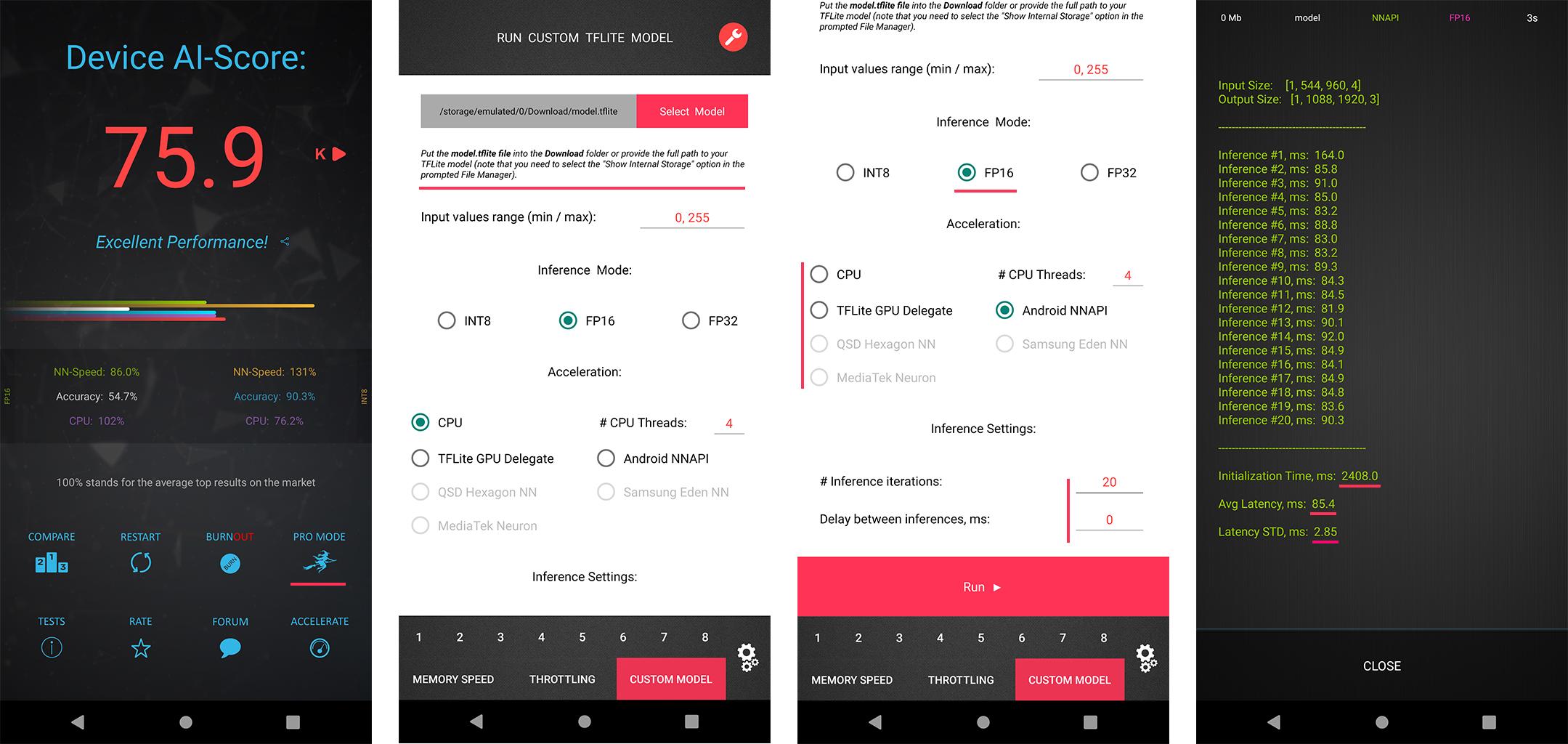}
}
\vspace{0.2cm}
\caption{Loading and running custom TensorFlow Lite models with AI Benchmark application. The currently supported acceleration options include Android NNAPI, TFLite GPU, Hexagon NN, Qualcomm QNN, MediaTek Neuron and Samsung ENN delegates as well as CPU inference through TFLite or XNNPACK backends. The latest app version can be downloaded at \url{https://ai-benchmark.com/download}}
\vspace{-1.2mm}
\label{fig:ai_benchmark_custom}
\end{figure*}

\section{Challenge}

In order to design an efficient and practical deep learning-based solution for the considered task that runs fast on mobile devices, one needs the following tools:

\begin{enumerate}
\item A large-scale high-quality dataset for training and evaluating the models. Real, not synthetically generated data should be used to ensure a high quality of the obtained model;
\item An efficient way to check the runtime and debug the model locally without any constraints as well as the ability to check the runtime on the target evaluation platform.
\end{enumerate}

This challenge addresses all the above issues. Real training data, tools, and runtime evaluation options provided to the challenge participants are described in the next sections.

\subsection{Dataset}

To get real and diverse data for the considered challenge, a novel dataset consisting of RGB-depth image pairs was collected using the ZED stereo camera\footnote{\url{https://www.stereolabs.com/zed/}} capable of shooting 2K images. It demonstrates an average depth estimation error of less than 0.2m for objects located closer than 8 meters~\cite{ortiz2018depth}, while more coarse predictions are also available for distances of up to 50 meters. Around 8.3K image pairs were collected in the wild over several weeks in a variety of places. For this challenge, the obtained images were downscaled to VGA resolution (640$\times$480 pixels) that is typically used on mobile devices for different depth-related tasks. The original RGB images were then considered as inputs, and the corresponding 16-bit depth maps~--- as targets. A sample RGB-depth image pair from the collected dataset is demonstrated in Fig.~\ref{fig:example_photos}.

\subsection{Local Runtime Evaluation}

When developing AI solutions for mobile devices, it is vital to be able to test the designed models and debug all emerging issues locally on available devices. For this, the participants were provided with the \textit{AI Benchmark} application~\cite{ignatov2018ai,ignatov2019ai} that allows to load any custom TensorFlow Lite model and run it on any Android device with all supported acceleration options. This tool contains the latest versions of \textit{Android NNAPI, TFLite GPU, Hexagon NN, Qualcomm QNN, MediaTek Neuron} and \textit{Samsung ENN} delegates, therefore supporting all current mobile platforms and providing the users with the ability to execute neural networks on smartphone NPUs, APUs, DSPs, GPUs and CPUs.

\smallskip

To load and run a custom TensorFlow Lite model, one needs to follow the next steps:

\begin{enumerate}
\setlength\itemsep{0mm}
\item Download AI Benchmark from the official website\footnote{\url{https://ai-benchmark.com/download}} or from the Google Play\footnote{\url{https://play.google.com/store/apps/details?id=org.benchmark.demo}} and run its standard tests.
\item After the end of the tests, enter the \textit{PRO Mode} and select the \textit{Custom Model} tab there.
\item Rename the exported TFLite model to \textit{model.tflite} and put it into the \textit{Download} folder of the device.
\item Select mode type \textit{(INT8, FP16, or FP32)}, the desired acceleration/inference options and run the model.
\end{enumerate}

\noindent These steps are also illustrated in Fig.~\ref{fig:ai_benchmark_custom}.

\subsection{Runtime Evaluation on the Target Platform}

In this challenge, we use the \textit{Raspberry Pi 4} single-board computer as our target runtime evaluation platform. It is based on the \textit{Broadcom BCM2711} chipset containing four Cortex-A72 ARM cores clocked at 1.5 GHz and demonstrates AI Benchmark scores comparable to entry-level Android smartphone SoCs~\cite{AIBenchmark202104}. The Raspberry Pi 4 supports the majority of Linux distributions, Windows 10 IoT build as well as Android operating system. In this competition, the runtime of all solutions was tested using the official TensorFlow Lite 2.5.0 Linux build~\cite{TensorFlowLite2021Linux} containing many important performance optimizations for the above chipset, the default \textit{Raspberry Pi OS} was installed on the device. Within the challenge, the participants were able to upload their TFLite models to our dedicated competition platform\footnote{\url{https://ml-competitions.com}} connected to a real Raspberry Pi 4 board and get instantaneous feedback: the runtime of their solution or an error log if the model contains some incompatible operations. The same setup was also used for the final runtime evaluation.

\begin{table*}[t!]
\centering
\resizebox{\linewidth}{!}
{
\begin{tabular}{l|c|cc|cccc|c|c}
\hline
Team \, & \, Author \, & \, Framework \, & \, Model Size, MB \, & \, si-RMSE$\downarrow$ \, & \, RMSE$\downarrow$ \, & \, LOG10$\downarrow$ \, & \, REL$\downarrow$ \, & \, Runtime, ms $\downarrow$ \, & \, Final Score \\
\hline
\hline
TCL & \, TCL \, & \, TensorFlow \, &  2.9 & \textBF{0.2773} & \textBF{3.47} & \textBF{0.1103} & 0.2997 & 46 & \textBF{298} \\
AIIA HIT & \, Zhenyu Li \, & \, PyTorch / TensorFlow \, &  1.5 & 0.311 & 3.79 & 0.1241 & 0.3427 & \textBF{37} & 232 \\
MiAIgo & \, ChaoMI \, & \, PyTorch / TensorFlow \, & 1.0 & 0.299 & 3.89 & 0.1349 & 0.3807 & 54 & 188 \\
Tencent GY-Lab & \, Parkzyzhang \, & \, PyTorch / TensorFlow \, & 3.4 & 0.303 & 3.8 & 0.1899 & 0.3014 & 68 & 141 \\
\hline
Tencent GY-Lab$^*$ & \, Parkzyzhang \, & \, PyTorch / TensorFlow \, & 3.4 & 0.2836 & 3.56 & 0.1121 & \textBF{0.2690} & 103 & 122 \\
\hline
SmartLab & \, RocheL \, & \, TensorFlow \, &  7.1 & 0.3296 & 4.06 & 0.1378 & 0.3662 & 65 & 102 \\
JMU-CVLab & \, mvc \, & \, PyTorch / TensorFlow \, &  3.5 & 0.3498 & 4.46 & 0.1402 & 0.3404 & 139 & 36 \\
ICL & \, Byung Hyun Lee \, & \, PyTorch / TensorFlow \, &  5.9 & 0.338 & 6.73 & 0.3323 & 0.5070 & 142 & 42 \\
\end{tabular}
}
\vspace{2.6mm}
\caption{\small{MAI 2022 Monocular Depth Estimation challenge results and final rankings. The runtime values were obtained on 640$\times$480 px images on the Raspberry Pi 4 device. Team \textit{TCL} is the challenge winner. $^*$ This model was the challenge winner in the previous MAI 2021 depth estimation challenge.}}
\label{tab:results}
\end{table*}

\subsection{Challenge Phases}

The challenge consisted of the following phases:

\vspace{-0.8mm}
\begin{enumerate}
\item[I.] \textit{Development:} the participants get access to the data and AI Benchmark app, and are able to train the models and evaluate their runtime locally;
\item[II.] \textit{Validation:} the participants can upload their models to the remote server to check the fidelity scores on the validation dataset, to get the runtime on the target platform, and to compare their results on the validation leaderboard;
\item[III.] \textit{Testing:} the participants submit their final results, codes, TensorFlow Lite models, and factsheets.
\end{enumerate}
\vspace{-0.8mm}

\subsection{Scoring System}

All solutions were evaluated using the following metrics:

\vspace{-0.8mm}
\begin{itemize}
\setlength\itemsep{-0.2mm}
\item Root Mean Squared Error (RMSE) measuring the absolute depth estimation accuracy,
\item Scale Invariant Root Mean Squared Error (si-RMSE) measuring the quality of relative depth estimation (relative position of the objects),
\item Average $\log_{10}$ and Relative (REL) errors~\cite{liu2015learning},
\item The runtime on the target Raspberry Pi 4 device.
\end{itemize}
\vspace{-0.8mm}

The score of each final submission was evaluated based on the next formula ($C$ is a constant normalization factor):

\smallskip
\begin{equation*}
\text{Final Score} \,=\, \frac{2^{-20 \cdot \text{si-RMSE}}}{C \cdot \text{runtime}},
\end{equation*}
\smallskip

During the final challenge phase, the participants did not have access to the test dataset. Instead, they had to submit their final TensorFlow Lite models that were subsequently used by the challenge organizers to check both the runtime and the fidelity results of each submission under identical conditions. This approach solved all the issues related to model overfitting, reproducibility of the results, and consistency of the obtained runtime/accuracy values.

\section{Challenge Results}

From above 70 registered participants, 7 teams entered the final phase and submitted valid results, TFLite models, codes, executables and factsheets. Table~\ref{tab:results} summarizes the final challenge results and reports si-RMSE, RMSE, LOG10 and REL measures and runtime numbers for each submitted solution on the final test dataset and on the target evaluation platform. The proposed methods are described in section~\ref{sec:solutions}, and the team members and affiliations are listed in Appendix~\ref{sec:apd:team}.

\subsection{Results and Discussion}

All solutions proposed in this challenge demonstrated a very high efficiency, being able to produce depth maps under 150 ms on the Raspberry 4 platform. As expected, all proposed architectures use a U-Net based structure, where MobileNets or EffientNets are generally used in the encoder part to extract relevant features from the input image. Team \textit{TCL} is the winner of the challenge: their model demonstrated the best accuracy results and a runtime of less than 50 milliseconds on the target device. Compared to the last year's winning challenge solution~\cite{zhang2021asimple}, this model is slightly more accurate while more than 2 times faster. The second best result was achieved by team \textit{AIIA HIT}. This solution is able to run at more than 27 FPS on the Raspberry Pi 4, thus demonstrating a nearly real-time performance, which is critical for many depth estimation applications. Overall, we can see a noticeable improvement in the efficiency of the proposed solutions compared to the models produced in the previous Mobile AI depth estimation challenge~\cite{ignatov2021fastDepth}, which allows for faster and more accurate depth estimation models on mobile devices.

\section{Challenge Methods}
\label{sec:solutions}

\begin{figure*}[h!]
\vspace{2mm}
\centering
\includegraphics[width=0.86\textwidth]{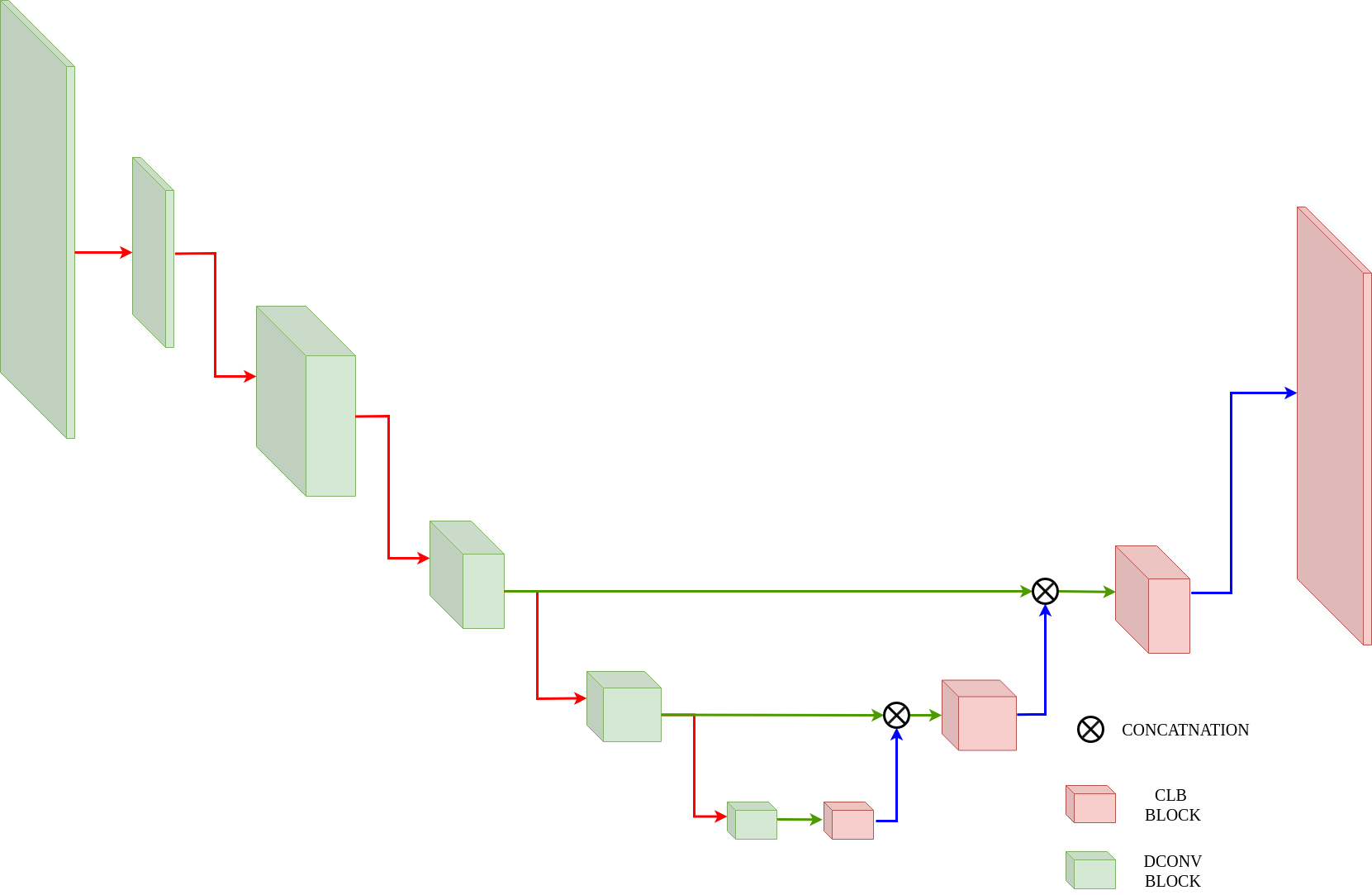}
\vspace{2mm}
\caption{Model architecture proposed by team TCL.}
\label{fig:TCL}
\end{figure*}

\noindent This section describes solutions submitted by all teams participating in the final stage of the MAI 2022 Monocular Depth Estimation challenge.

\subsection{TCL}

Team TCL proposed a UNet-like~\cite{ronneberger2015u} architecture presented in Fig.~\ref{fig:TCL}. The input image is first resized from 640$\times$480 to 160$\times$128 pixels and then passed to a simplified MobileNetV3-based~\cite{howard2019searching} encoder. The lowest resolution blocks were completely remove from the encoder module, which had a little impact on the overall quality, but led to a noticeable latency drop. The output of the encoder is then passed to the decoder module that contains only Collapsible Linear Blocks (CLB)~\cite{MLSYS2022_ac627ab1}. Finally, the output of the decoder is resized with a scale factor of 10 from 48$\times$64 to 480$\times$640 pixels.

The objective function described in \cite{arxiv.1907.10326} was used as it provided better results
than the standard RMSE loss. Model parameters were optimized using the Adam~\cite{kingma2014adam} algorithm with the cosine learning rate scheduler.

\subsection{AIIA HIT}

\begin{figure*}[h!]
\vspace{2mm}
\centering
\includegraphics[width=0.86\textwidth]{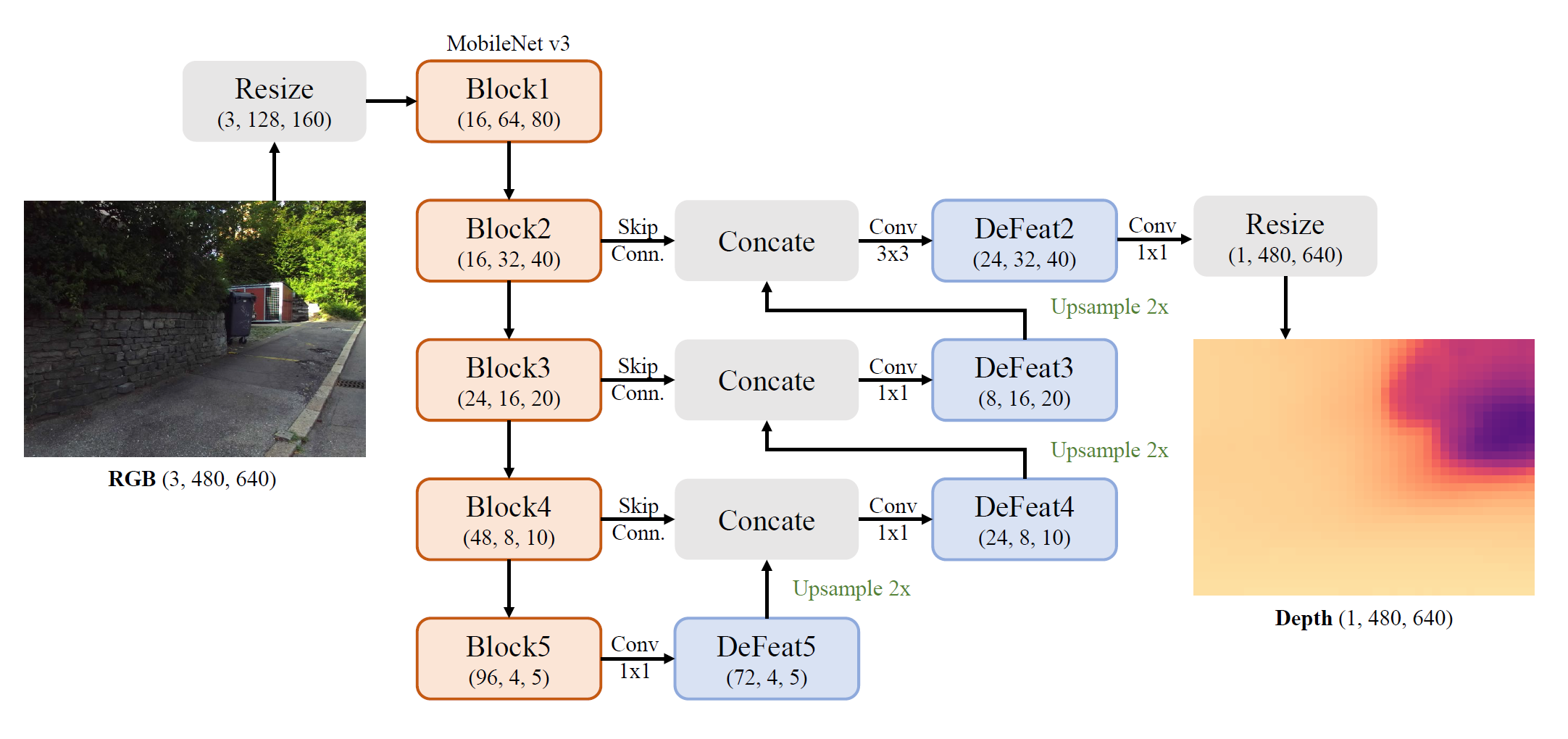}
\vspace{2mm}
\caption{An overview of the model proposed by team AIIA HIT.}
\label{fig:AIIA_HIT}
\end{figure*}

Team AIIA HIT followed a similar approach of using a UNet-like architecture with a MobileNetV3-based encoder and an extremely lightweight decoder part that only contains a single $3\times3$ convolution and four $1\times1$ convolution modules. The architecture of this solution~\cite{li2022litedepth} is demonstrated in Fig.~\ref{fig:AIIA_HIT}. The training process included a custom data augmentation strategy called the R$^{2}$ crop (Random locations and Random size changes of patches). Four different loss functions were used for training. The first SILog~\cite{eigen2014depth} loss function was defined as:

\begin{equation}
  \mathcal{L}_{silog} = \alpha \sqrt{\frac{1}{N}\sum\limits_i^N e_{i}^{2} - \frac{\lambda}{N^2}(\sum\limits_i^N e_i)^2},
  \label{eq:silog_loss}
\end{equation}
where $e_i = \log \hat{d}_i - \log d_i$ is the log difference between the ground truth $d_i$ and the predicted $\hat{d}_i$ depth maps. $N$ here denotes the number of pixels having valid ground truth values. The other three losses are the gradient loss $\mathcal{L}_{grad}$, the virtual norm loss $\mathcal{L}_{vnl}$~\cite{yin2019enforcing}, and the robust loss $\mathcal{L}_{robust}$~\cite{barron2019general} specified as follows:

\begin{equation}
  \mathcal{L}_{grad} = \frac{1}{T}\sum_i\left(\left\|\nabla_x\hat{d}_i - \nabla_x d_i\right\|_1 + \left\|\nabla_y\hat{d}_i - \nabla_y d_i\right\|_1\right)
  \label{eq:grad_loss}
\end{equation}
\begin{equation}
  \mathcal{L}_{vnl} = \sum\limits_i^N\left\|\hat{n}_i - n_i \right\|_1,
  \label{eq:vnl_loss}
\end{equation}
\begin{equation}
  \mathcal{L}_{robust} = \frac{|\alpha - 2 |}{\alpha}\left(\left(\frac{(x/c)^2}{|\alpha - 2|} \right)^{\alpha/2}-1 \right),
  \label{eq:robust_loss}
\end{equation}
where $\nabla$ is the gradient operator,  $\alpha=1$ and $c=2$. It should be noted that the $\mathcal{L}_{vnl}$ loos slightly differs from the original implementation as the points here are sampled from the reconstructed point clouds (instead of the ground truth maps) to filter invalid samples as this helped the model to converge at the beginning of the training. The final loss function was defined as follows:

\begin{equation}
  \mathcal{L}_{depth} = w_1\mathcal{L}_{silog} + w_2\mathcal{L}_{grad} + w_3\mathcal{L}_{vnl} + w_4\mathcal{L}_{robust}.
\end{equation}
where $w_1=1$, $w_2=0.25$, $w_3=2.5$, and $w_4=0.6$. To achieve a better performance, a structure-aware distillation strategy~\cite{liu2020structured} was applied, where the Swin Transformer trained with the $\mathcal{L}_{depth}$ loss as a teacher model. During the distillation, multi-level distilling losses were adopted to provide a supervision on immediate features. The final model was trained in two stages. First, the model was trained with the $\mathcal{L}_{depth}$ only. During the second state, the model was fine-tuned with a combination of the distillation and depth loss functions:
\begin{equation}
  \mathcal{L_{\text{stage-2}}} = \mathcal{L}_{depth} + w\mathcal{L}_{distill}
\end{equation}
where $w=10$.

\subsection{MiAIgo}

\begin{figure*}[h!]
\vspace{2mm}
\centering
\includegraphics[width=0.86\textwidth]{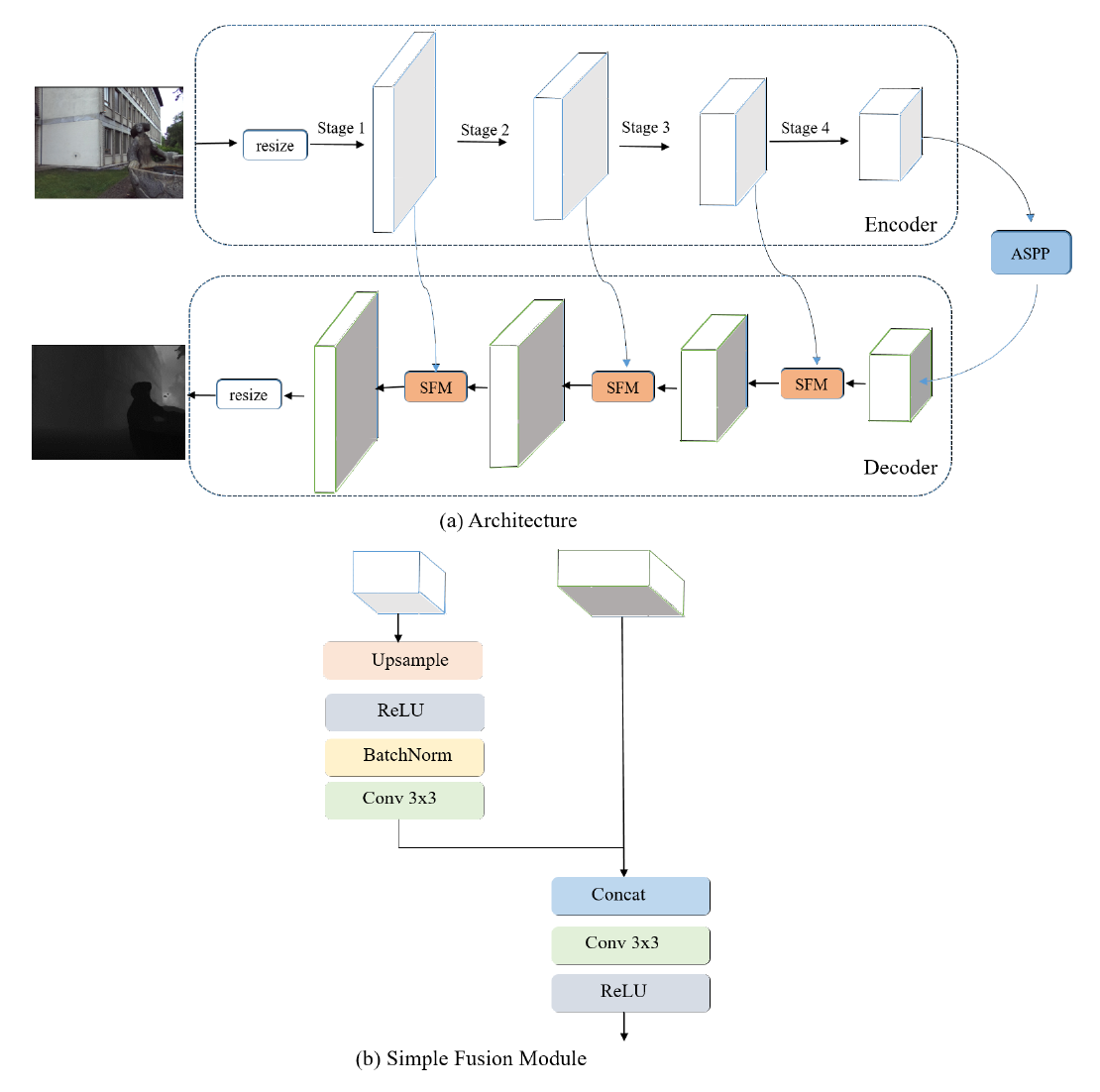}
\vspace{2mm}
\caption{The architecture of the model and the structure of the Simple Fusion Module (SFM) proposed by team MiAIgo.}
\label{fig:MiAIgo}
\end{figure*}

Team MiAIgo used a modified GhostNet~\cite{arxiv.1911.11907} architecture with a reduced number of channels for feature extraction. The overall architecture of the proposed model is demonstrated in Fig.~\ref{fig:MiAIgo}. The input image is first resized from 640${\times}$480 to 128${\times}$96 pixels and then passed to the encoder module consisting of four blocks. The Atrous Spatial Pyramid Pooling (ASPP)~\cite{arxiv.1706.05587} module was placed on the top of the encoder to process multi-scale contextual information and increase the receptive field. According to the experiments, ASPP provides a better accuracy with minor inference time expenses. Besides that, the Simple Fusion Module (SFM) was designed and used in the decoder part of the model. The SFM concatenates the outputs of different stages with the decoder feature maps to achieve better fidelity results. An additional nearest neighbor resizing layer was placed after the decoder part to upscale the model output to the target resolution. The model was trained to minimize the SSIM loss function, its parameters were optimized using the Adam algorithm with a learning rate of 8e--4 and a batch size of 8.

\subsection{Tencent GY-Lab}

\begin{figure*}[h!]
\vspace{2mm}
\centering
\includegraphics[width=0.86\textwidth]{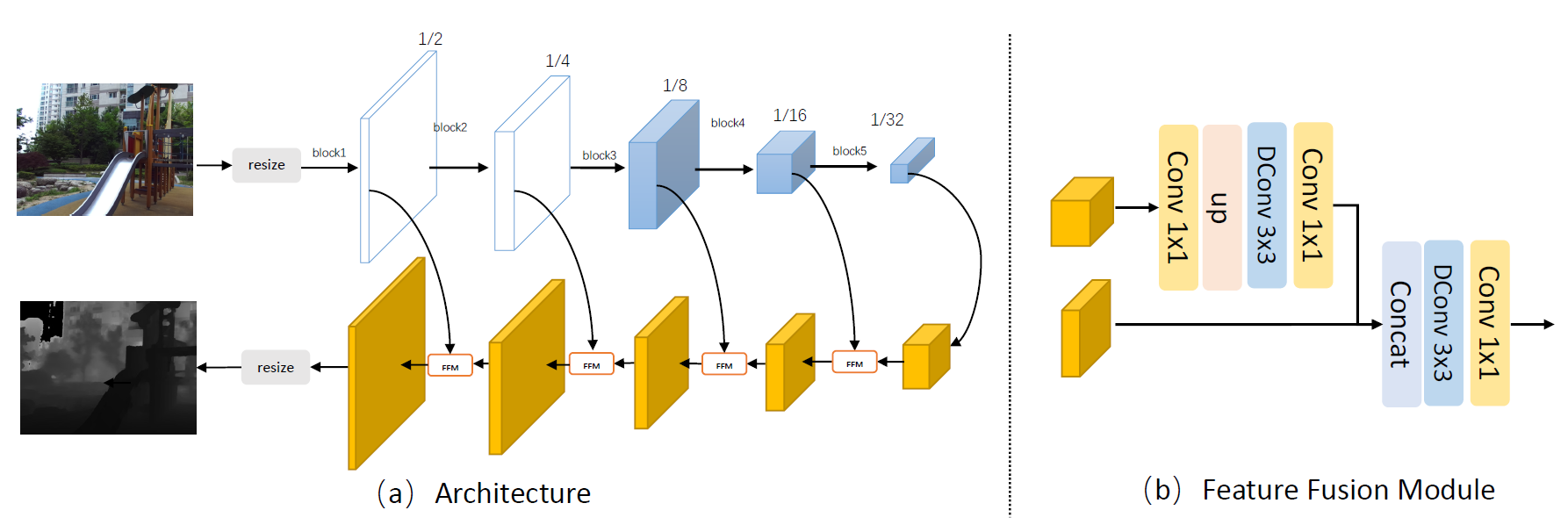}
\vspace{2mm}
\caption{The model architecture and the structure of the Feature Fusion Module (FFM) proposed by team Tencent GY-Lab.}
\label{fig:light_architecture}
\end{figure*}

Team Tencent GY-Lab proposed a U-Net like architecture presented in Fig.~\ref{fig:light_architecture}, where a MobileNet-V3~\cite{howard2019searching} based encoder is used for dense feature extraction. To reduce the amount of computations, the input image is first resized from 640$\times$480 to 128$\times$96 pixels and then passed to the encoder module consisting of five blocks. The outputs of each block are processed by the Feature Fusion Module (FFM) that concatenates them with the decoder feature maps to get better fidelity results. The authors use one additional \textit{nearest neighbor} resizing layer on top of the model to upscale the output to the target resolution. Knowledge distillation~\cite{hinton2015distilling} is further used to improve the quality of the reconstructed depth maps: a bigger ViT-Large~\cite{dosovitskiy2020image} was first trained on the same dataset and then its features obtained before the last activation function were used to guide the smaller network. This process allowed to decrease the si-RMSE score from $0.3304$ to $0.3141$. The proposed model was therefore trained to minimize a combination of the distillation loss (computed as $L_2$ norm between its features from the last convolutional layer and the above mentioned features from the larger model), and the depth estimation loss proposed in~\cite{lee2019big}. The network parameters were optimized for 500 epochs using Adam~\cite{kingma2014adam} with a learning rate of $8e-3$ and a polynomial decay with a power of $0.9$. The model was implemented and trained with PyTorch and then converted to TensorFlow Lite using ONNX as an intermediate representation. A more detailed description of the proposed solution is provided in~\cite{zhang2021asimple}.

\subsection{SmartLab}

\begin{figure*}[h!]
\vspace{2mm}
\centering
\includegraphics[width=0.86\textwidth]{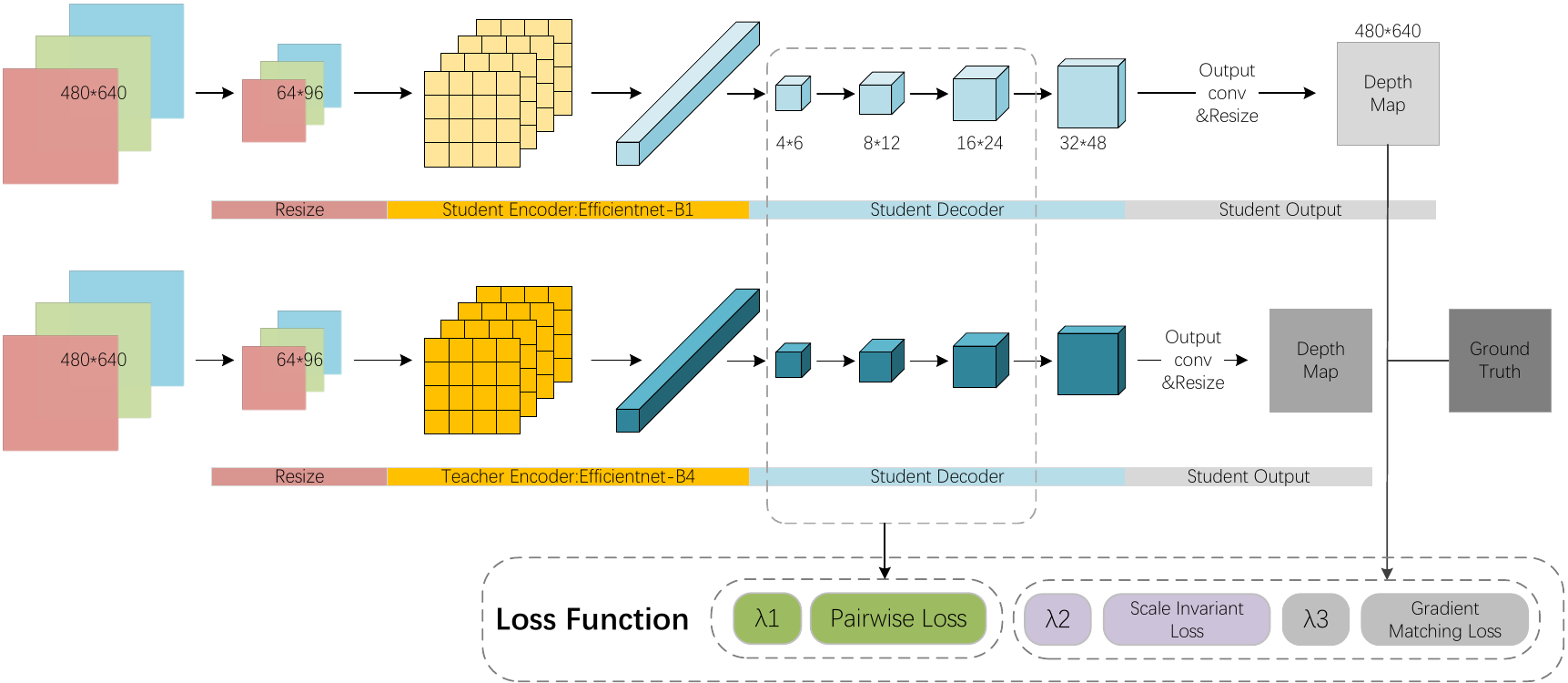} \vspace{4mm} \\
\includegraphics[width=0.36\textwidth]{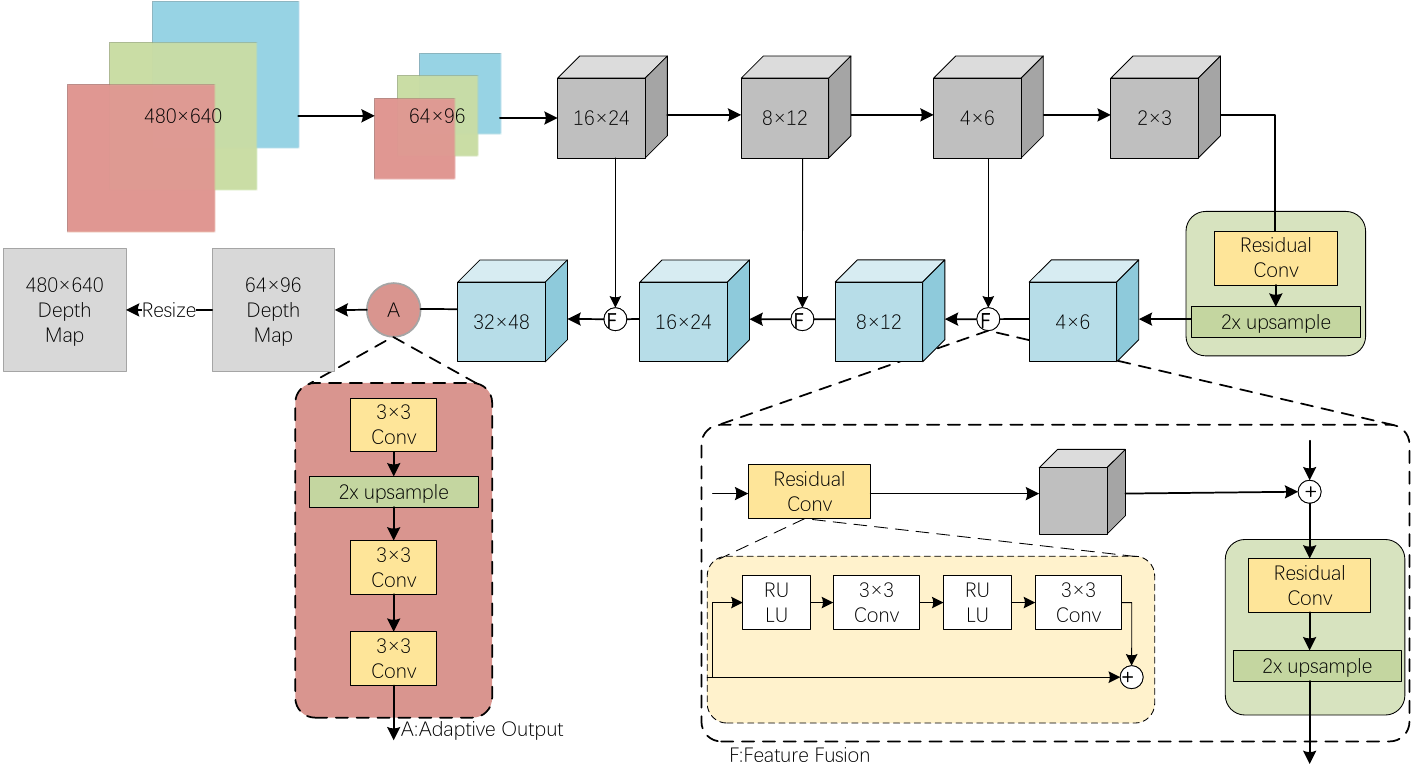} \hspace{4mm}
\includegraphics[width=0.44\textwidth]{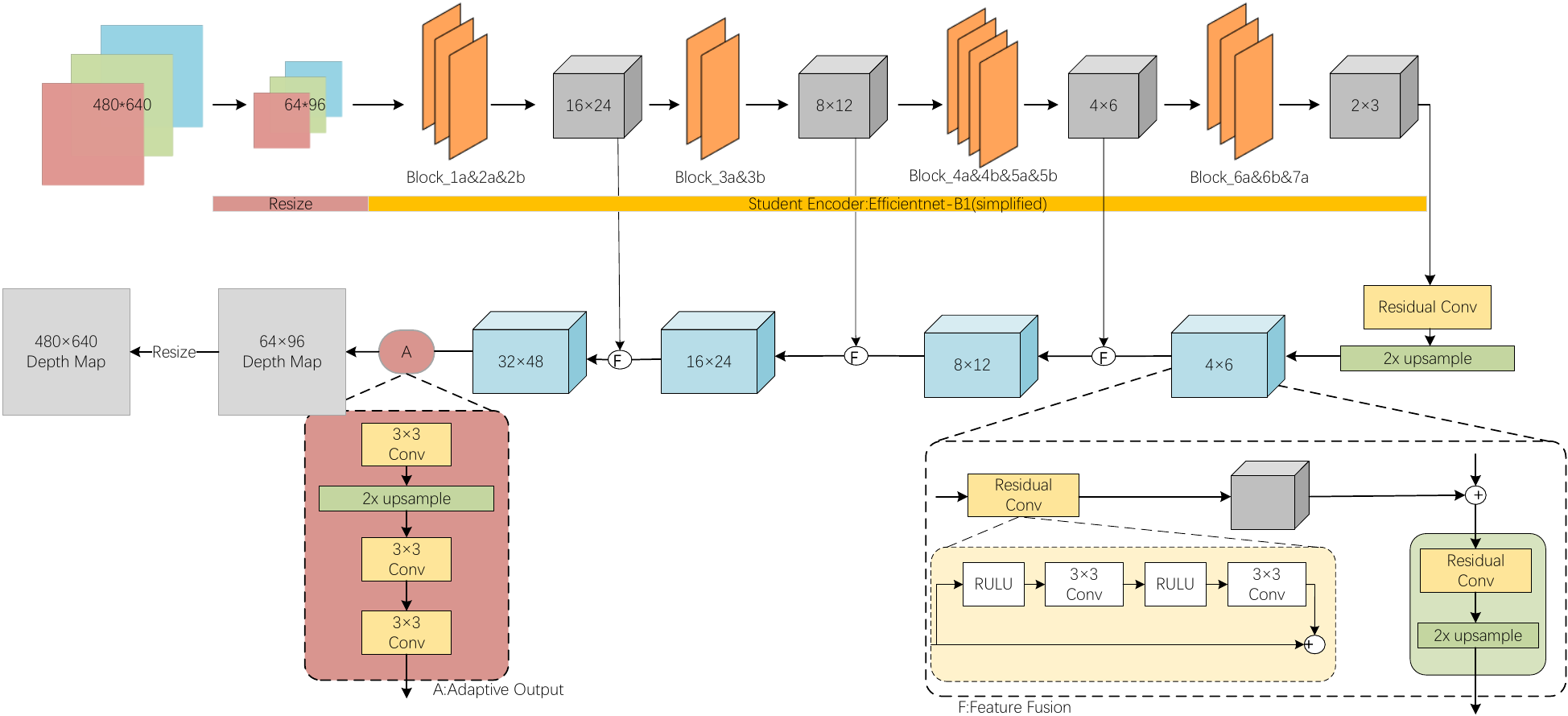}
\vspace{2mm}
\caption{An overview of the training strategy (top) and the architecture of the teacher (bottom left) and student (bottom right) models proposed by team SmartLab.}
\label{fig:SmartLab}
\end{figure*}

The solution proposed by team SmartLab largely relies on transfer learning and consists of two networks as shown in Fig.~\ref{fig:SmartLab}. The representation ability of the pretrained teacher network is transferred to the student network via the pairwise distillation~\cite{2019structuredliu}. Scale-invariant loss~\cite{2014deptheigen} and gradient matching loss~\cite{2020towardsranftl} were used to train the student network, guiding the model to predict more accurate depth maps. Both teacher and student networks are based on the encoder-decoder architecture with the EfficientNet used as a backbone. The network structure design mainly refers to the MiDaS~\cite{2020towardsranftl} paper, the encoder uses the EfficientNet-B1 architecture, and the basic block of the decoder is the Feature Fusion Module (FFM).

The model was trained using three loss functions: the scale invariant loss~\cite{2014deptheigen} to measure the discrepancy between the output of the student network and the ground truth depth map; the scale-invariant gradient matching loss~\cite{2020towardsranftl}; and the pairwise distillation loss~\cite{2019structuredliu} to force the student network to produce similar feature maps as the outputs of the corresponding layers of the teacher network. The latter loss is defined as:

\begin{equation}
{\mathcal{L}_{pa}}(S,T)=\frac{1}{w \times h} \sum_{i} \sum_{j} (a^s_{ij}-a^t_{ij})^2,
\label{e4}
\end{equation}
\begin{equation}
a_{ij}=f_i^Tf_j/({\vert\vert f_i \vert\vert}_2 \times {\vert\vert f_j \vert\vert}_2 ),
\label{e3}
\end{equation}

\noindent where $F_t \in \mathbb{R}^{h\times w \times c_1} $ and $F_s \in \mathbb{R}^{h\times w \times c_2} $ are the feature maps from the teacher and student networks, respectively, and $f$ denotes one row of a feature map ($F_t$ or $F_s$). Model parameters were optimized using the Adam algorithm for 100 epochs with a batch size of 2, input images were cropped to patches of size $64\times96$ and flipped randomly horizontally for data augmentation.

\subsection{JMU-CVLab}

\begin{figure*}[h!]
\vspace{2mm}
\centering
\includegraphics[width=0.86\textwidth]{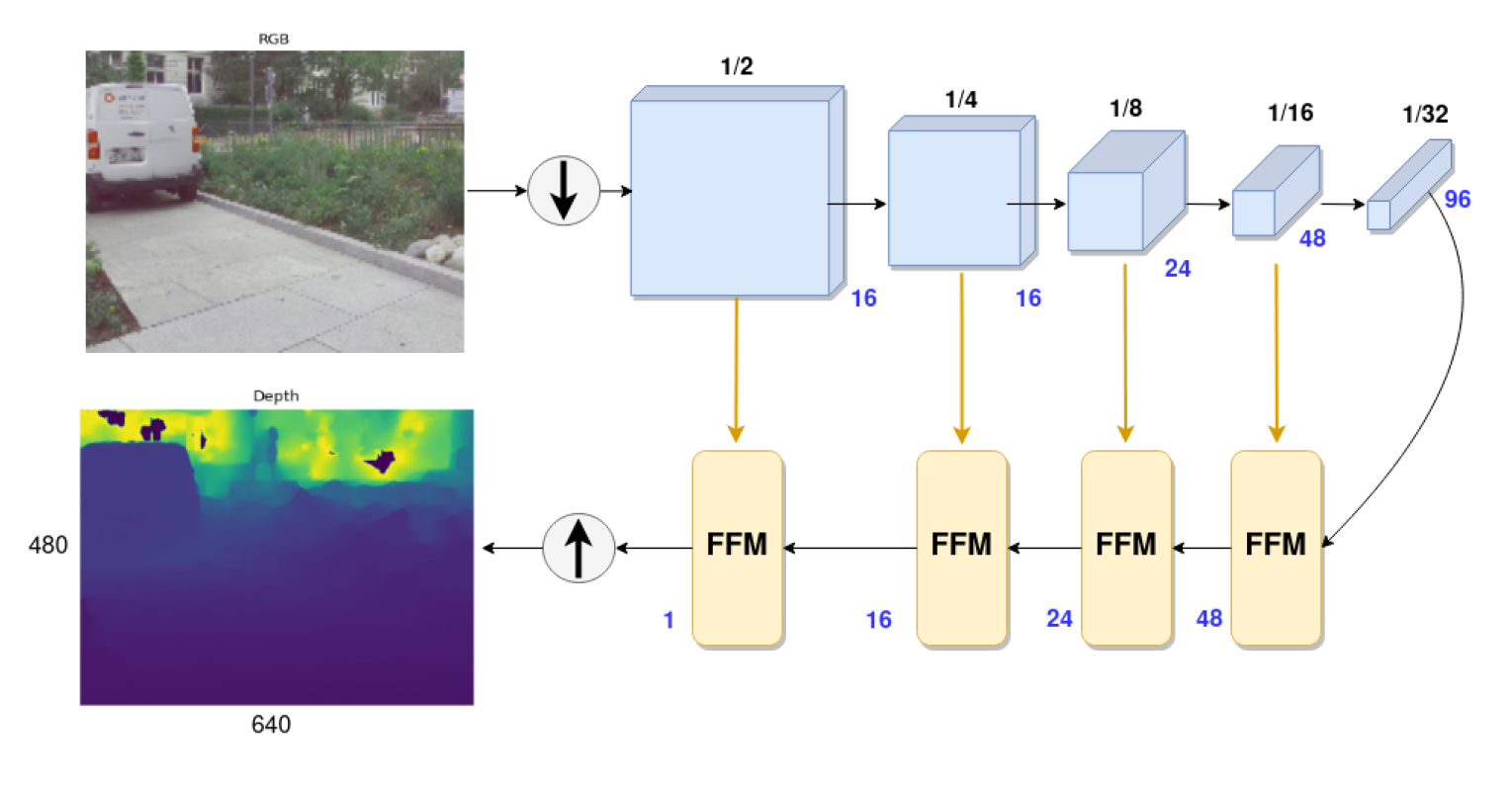}
\vspace{2mm}
\caption{Model architecture proposed by team JMU-CVLab.}
\label{fig:JMU}
\end{figure*}

The solution developed by team JMU-CVLab was inspired by the FastDepth~\cite{arxiv.1903.03273} paper and the last year's winning challenge solution~\cite{9523055}, its architecture is shown in Fig.~\ref{fig:JMU}. The input image is first resized to 160$\times$128 pixels and then passed to a MobileNet-V3 backbone pre-trained on the ImageNet dataset that is used for feature extraction. The obtained features are then processed with a series of decoder FFM blocks~\cite{9523099} that perform an upsampling operation and reduce the number of channels by projecting them to a lower space using separable and pointwise convolutions. The model was trained with a combination of the L2, SSIM and si-RMSE loss functions. Model parameters were optimized with the Adam algorithm for 100 epochs using a learning rate 0.0001 and batch size 64. Basic data augmentations were used during the training including horizontal flips, random rotations limited by a 15$^{\circ}$ angle, color and brightness changes.

\subsection{ICL}

\begin{figure*}[h!]
\vspace{2mm}
\centering
\includegraphics[width=0.86\textwidth]{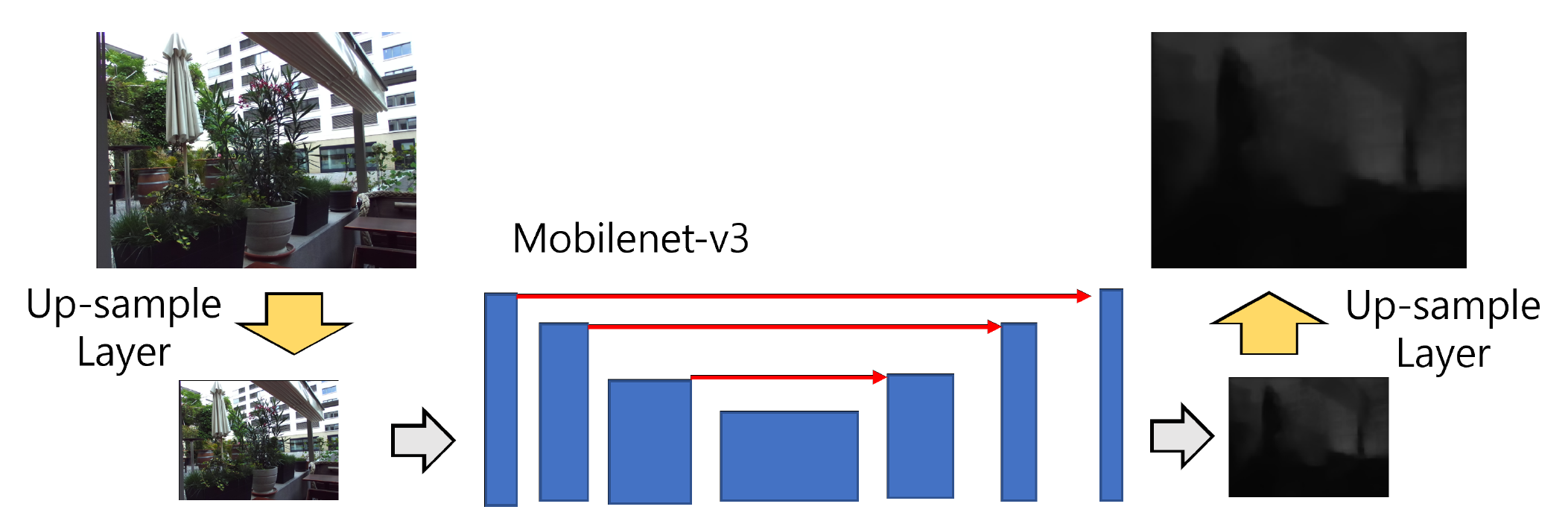}
\vspace{2mm}
\caption{ICL.}
\label{fig:ICL}
\end{figure*}

Team ICL proposed a solution that also uses the MobileNet-V3 network as a feature extractor (Fig.~\ref{fig:ICL}). The authors first fine-tuned the ViT model pre-trained on the KITTI depth estimation dataset~\cite{Uhrig2017THREEDV}, and then used it as a teacher network. Next, the proposed student model was first pre-trained for the semantic segmentation task and then fine-tuned on depth data. The student network was trained using a combination of the si-RMSE, gradient and knowledge distillation losses. Model parameters were optimized using the AdamW~\cite{loshchilov2017decoupled} optimizer with a batch size of 32 and the cosine learning rate scheduler.

\section{Additional Literature}

An overview of the past challenges on mobile-related tasks together with the proposed solutions can be found in the following papers:

\begin{itemize}
\item Learned End-to-End ISP:\, \cite{ignatov2019aim,ignatov2020aim,ignatov2022microisp,ignatov2022pynetv2}
\item Perceptual Image Enhancement:\, \cite{ignatov2018pirm,ignatov2019ntire}
\item Image Super-Resolution:\, \cite{ignatov2018pirm,lugmayr2020ntire,cai2019ntire,timofte2018ntire}
\item Bokeh Effect Rendering:\, \cite{ignatov2019aimBokeh,ignatov2020aimBokeh}
\item Image Denoising:\, \cite{abdelhamed2020ntire,abdelhamed2019ntire}
\end{itemize}

\section*{Acknowledgements}

We thank the sponsors of the Mobile AI and AIM 2022 workshops and challenges: AI Witchlabs, MediaTek, Huawei, Reality Labs, OPPO, Synaptics, Raspberry Pi, ETH Z\"urich (Computer Vision Lab) and University of W\"urzburg (Computer Vision Lab).

\appendix
\section{Teams and Affiliations}
\label{sec:apd:team}

\bigskip

\subsection*{Mobile AI 2022 Team}
\noindent\textit{\textbf{Title: }}\\ Mobile AI 2022 Challenge on Single-Image Depth Estimation on Mobile Devices\\
\noindent\textit{\textbf{Members:}}\\ Andrey Ignatov$^{1,2}$ \textit{(andrey@vision.ee.ethz.ch)}, Grigory Malivenko \textit{(grigory.malivenko @gmail.com)}, Radu Timofte$^{1,2,3}$ \textit{(radu.timofte@vision.ee.ethz.ch)}\\
\noindent\textit{\textbf{Affiliations: }}\\
$^1$ Computer Vision Lab, ETH Zurich, Switzerland\\
$^2$ AI Witchlabs, Switzerland\\
$^3$ University of Wuerzburg, Germany\\

\subsection*{TCL}
\noindent\textit{\textbf{Title:}}\\ Simplified UNET Architecture For Fast Depth Estimation on Edge Devices \\
\noindent\textit{\textbf{Members: }}\\ \textit{Lukasz Treszczotko (lukasz.treszczotko@tcl.com)}, Xin Chang, Piotr Ksiazek, Michal Lopuszynski, Maciej Pioro, Rafal Rudnicki, Maciej Smyl, Yujie Ma \\
\noindent\textit{\textbf{Affiliations: }}\\
TCL Research Europe, Warsaw, Poland\\

\subsection*{AIIA HIT}
\noindent\textit{\textbf{Title:}}\\ Towards Fast and Accurate Depth Estimation on Mobile Devices~\cite{li2022litedepth} \\
\noindent\textit{\textbf{Members: }}\\ \textit{ Zhenyu Li (zhenyuli17@hit.edu.cn)}, Zehui Chen, Jialei Xu, Xianming Liu, Junjun Jiang \\
\noindent\textit{\textbf{Affiliations: }}\\
Harbin Institute of Technology, China \\

\subsection*{MiAIgo}
\noindent\textit{\textbf{Title:}}\\ SL-Depth : A Superior Lightweight Model for Monocular Depth Estimation on Mobile Devices \\
\noindent\textit{\textbf{Members: }}\\ \textit{ XueChao Shi (shixuechao@xiaomi.com)}, Difan Xu, Yanan Li, Xiaotao Wang, Lei Lei \\
\noindent\textit{\textbf{Affiliations: }}\\
Xiaomi Inc., China \\

\subsection*{Tencent GY-Lab}
\noindent\textit{\textbf{Title:}}\\ A Simple Baseline for Fast and Accurate Depth Estimation on Mobile Devices~\cite{zhang2021asimple} \\
\noindent\textit{\textbf{Members: }}\\ \textit{Ziyu Zhang (parkzyzhang@tencent.com)}, Yicheng Wang, Zilong Huang, Guozhong Luo, Gang Yu, Bin Fu \\
\noindent\textit{\textbf{Affiliations: }}\\
Tencent GY-Lab, China\\

\subsection*{SmartLab}
\noindent\textit{\textbf{Title:}}\\ Fast and Accurate Monocular Depth Estimation via Knowledge Distillation \\
\noindent\textit{\textbf{Members: }}\\ \textit{ Jiaqi Li (lijiaqi\_mail@hust.edu.cn)}, Yiran Wang, Zihao Huang, Zhiguo Cao \\
\noindent\textit{\textbf{Affiliations: }}\\
National Key Laboratory of Science and Technology on Multi-Spectral Information Processing, School of Artificial Intelligence and Automation, Huazhong University of Science and Technology, China\\

\subsection*{JMU-CVLab}
\noindent\textit{\textbf{Title:}}\\ MobileNetV3 FastDepth \\
\noindent\textit{\textbf{Members: }}\\ \textit{ Marcos V. Conde (marcos.conde-osorio@uni-wuerzburg.de)}, Denis Sapozhnikov \\
\noindent\textit{\textbf{Affiliations: }}\\
University of Wuerzburg, Germany\\

\subsection*{ICL}
\noindent\textit{\textbf{Title:}}\\ Monocular Depth Estimation Using a Simple Network \\
\noindent\textit{\textbf{Members: }}\\ \textit{ Byeong Hyun Lee (ldlqudgus756@snu.ac.kr)}, Dongwon Park, Seongmin Hong, Joonhee Lee, Seunggyu Lee, Se Young Chun \\
\noindent\textit{\textbf{Affiliations: }}\\
Department of Electrical and Computer Engineering, Seoul National University, South Korea \\

{\small
\bibliographystyle{splncs04}
\bibliography{egbib}
}

\end{document}